%% file: main.tex
\documentclass{article}

\input{prestuff}

\usepackage[preprint]{neurips_2025}

\usepackage[utf8]{inputenc} 
\usepackage[T1]{fontenc}    
\usepackage{hyperref}       
\usepackage{url}            
\usepackage{booktabs}       
\usepackage{amsfonts}       
\usepackage{nicefrac}       
\usepackage{microtype}      
\usepackage{xcolor}         
\usepackage{algorithmic}
\usepackage{verbatim}
\usepackage{xspace,enumitem}
\newcommand{\sys}{{\sc Sims}\xspace}
\newcommand{\syspr}{{\sc Sims-Pr}\xspace}
\newcommand{\syscs}{{\sc Sims-Cs}\xspace}

\title{Self-Improving Model Steering}

\author{Rongyi Zhu, Yuhui Wang, Tanqiu Jiang, Jiacheng Liang, Ting Wang \\
  Stony Brook University  \\
  \texttt{\{rozzhu, wang155, tanjiang, jiachliang, twang\}@cs.stonybrook.edu}
}

\begin{document}

\maketitle

\input{abstract}
\input{intro}

\input{relwk}

\input{prelim}
\input{method}

\input{exps}

\input{conclude}

\bibliographystyle{plain}
\bibliography{main}

\input{app}


\end{document}

%% file: prestuff.tex
\usepackage{epsfig,amsmath,amsfonts,epsfig,multirow,makecell,caption,soul,csquotes,color,wrapfig,subcaption,mathtools,bm,spverbatim,booktabs,diagbox}
\usepackage[e]{esvect}
\usepackage[most]{tcolorbox}

\makeatletter
\newif\if@restonecol
\makeatother

\usepackage[boxed, ruled, vlined, linesnumbered]{algorithm2e}
\SetKwRepeat{Do}{do}{while}

\newenvironment{changemargin}[2]{\begin{list}{}{
	\setlength{\topsep}{0pt}\setlength{\leftmargin}{0pt}
	\setlength{\rightmargin}{0pt}
	\setlength{\listparindent}{\parindent}
	\setlength{\itemindent}{\parindent}
	\setlength{\parsep}{0pt plus 1pt}
	\addtolength{\leftmargin}{#1}\addtolength{\rightmargin}{#2}
	}\item}
	{\end{list}}

\usepackage[first=0,last=9]{lcg}
\usepackage{colortbl}
\definecolor{Gray}{gray}{0.8}
\colorlet{Red}{red!10!white}
\colorlet{Blue}{blue!10!white}

\usepackage{hyperref}

\newcommand{\msec}[1]{\S\ref{#1}}

\newcommand{\meq}[1]{Eq.\,\ref{#1}}
\newcommand{\mcite}[1]{\,\cite{#1}}

\newcommand{\mct}[1]{{\em #1})}

\newtcolorbox{mtbox}[1]{left=0.25mm, right=0.25mm, top=0.25mm, bottom=0.25mm, sharp corners, colframe=white!50!white, boxrule=0.5pt, title={#1}, fonttitle=\bfseries, coltitle=red!50!black,
breakable}

\usepackage{scalerel}[2016/12/29]

\makeatletter
\providecommand{\leadsfrom}{%
  \mathrel{\mathpalette\reflect@squig\relax}%
}
\newcommand{\reflect@squig}[2]{%
  \reflectbox{$\m@th#1\leadsto$}%
}
\makeatother

\def\eqref#1{equation~\ref{#1}}

\def\1{\bm{1}}

\DeclareMathAlphabet{\mathsfit}{\encodingdefault}{\sfdefault}{m}{sl}
\SetMathAlphabet{\mathsfit}{bold}{\encodingdefault}{\sfdefault}{bx}{n}

\def\gA{{\mathcal{A}}}

\def\gD{{\mathcal{D}}}

\def\gH{{\mathcal{H}}}

\def\gM{{\mathcal{M}}}

\def\gO{{\mathcal{O}}}



%% file: abstract.tex
\begin{abstract}
Model steering represents a powerful technique that dynamically aligns large language models (LLMs) with human preferences during inference. However, conventional model-steering methods rely heavily on externally annotated data, not only limiting their adaptability to varying contexts but also tethering their effectiveness to annotation quality. In this paper, we present \sys, the first self-improving model-steering framework that operates without relying on external supervision. At its core, \sys autonomously generates and refines contrastive samples through iterative self-improvement cycles, enabling adaptive, context-specific steering. Additionally, \sys employs novel strategies, including prompt ranking and contrast sampling, to further enhance steering efficacy. Extensive evaluation across diverse LLMs and benchmarks demonstrates that \sys substantially outperforms existing methods in steering effectiveness and adaptability, highlighting self-improving model steering as a promising direction for future research on inference-time LLM alignment. 
\end{abstract}

%% file: intro.tex
\section{Introduction}
\label{sec:intro}

\begin{wrapfigure}{r}{0.52\textwidth}
    \begin{center}
    \vspace{-5pt}
    \includegraphics[width=0.52\textwidth]{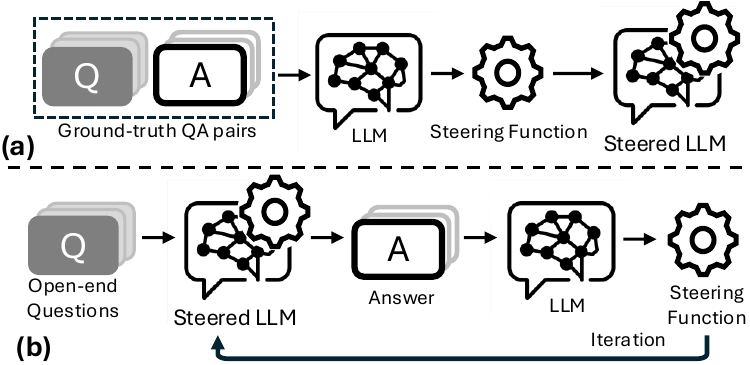}
    \end{center}
    \caption{\small Comparison of (a) conventional and (b) self-improving model steering.}
    \label{fig:teaser}
\end{wrapfigure}
In contrast to pre- or post-training alignment of large language models (LLMs)\mcite{rlhf,lee2023rlaif}, model steering\mcite{panickssery2023steering,li2023inference,qiu2024spectral} represents a compelling alternative with distinct advantages. By dynamically modifying latent activations during inference, steering approaches enable adaptive, context-sensitive alignment that can respond to specific values and intentions without requiring complete model retraining. A variety of model steering methods have been proposed, ranging from linear transformation and projection\mcite{panickssery2023steering,li2023inference} to subspace learning and optimization\mcite{qiu2024spectral,pham2024householder,zhang2024truthx,cao2024personalized}. However, as illustrated in Figure~\ref{fig:teaser}\,(a), most existing methods rely heavily on externally annotated data for steering direction optimization, creating multiple critical limitations. This dependence not only restricts adaptability to varying contexts but also directly tethers steering effectiveness to annotation quality. Furthermore, the static nature of annotated data also leads to diminishing returns as sample sizes increase; performance improvement typically plateaus beyond a certain threshold, rendering additional data collection increasingly inefficient despite its substantial costs.

To overcome these limitations, in this paper, we present \sys,\footnote{\sys: \ul{S}elf-\ul{I}mproving \ul{M}odel \ul{S}teering.} a novel self-improving model-steering framework. As illustrated in Figure~\ref{fig:teaser}\,(b), \sys distinguishes itself from conventional methods through two fundamental innovations. \mct{i} \ul{Supervision-free steering} -- \sys eliminates dependency on external data annotation by leveraging self-generated contrastive samples to derive steering directions. This paradigm shift substantially enhances adaptability to varying contexts and data distributions. \mct{ii} \ul{Iterative self-improvement} -- through cycles of self-evaluation and regeneration, \sys continuously refines steering directions to more effectively differentiate desirable and undesirable behaviors, leading to consistent performance gains across iterations. Additionally, \sys introduces two novel strategies to further enhance steering efficacy, including: {\em prompt ranking} that leverages the model's own judgment to generate preference signals and {\em contrast sampling} that selects the informative samples for refining steering directions.

Through extensive evaluation across diverse LLMs and benchmarks, we show that \sys effectively steers LLMs towards desirable behaviors, outperforming or matching existing model-steering methods that rely on externally annotated data. For instance, \sys improves the length-controlled WinRate of {\tt llama3-8b} on Alpaca-Eval\mcite{dubois2025lengthcontrolledalpacaevalsimpleway} from 2.86 to 11.89 in just one iteration. Our ablation study further reveals that \sys steadily enhances steering effectiveness across iterations, while prompt ranking and contrastive sampling substantially improve steering efficiency. For example, \sys increases the Arena-Hard\mcite{li2024crowdsourced} score sharply from 15.3 to 33.4 from the first iteration to the third iteration. The findings highlight self-improving model steering as a promising direction for future research on inference-time preference optimization.

Our contributions can be summarized as follows. 
\begin{itemize}[leftmargin=*]
\item To the best of our knowledge, this represents the first work on model steering without external supervision, exploring a new paradigm for inference-time LLM alignment.
\item We introduce \sys, a novel self-improving model-steering framework that iteratively refines steering directions through self-improvement cycles, enabling adaptive, context-specific
steering.
\item We conduct an extensive evaluation to validate that \sys effectively guides LLMs towards desirable behaviors, consistently outperforming or matching existing methods that require externally annotated data.
\end{itemize}

%% file: relwk.tex
\section{Related Work}
\label{sec:relwk}

{\bf Model steering.} Compared with pre- or post-training alignment of LLMs\mcite{rlhf,lee2023rlaif}, model steering represents an attractive alternative that dynamically modifies latent activations during inference\mcite{turner2023steering, liu2023context, zou2023representation, wu2024reft, chalnev2411improving, lee2024programming, he2024context, fang2024alphaedit, rodriguez2024controlling, wang2024semantics, liu2024reducing, cao2024personalized}. Existing model-steering methods can be categorized based on how the steering vectors are derived. One line of work\mcite{turner2023steering,panickssery2023steering,li2023inference} defines steering vectors as linear transformations or projections derived from model activations. For instance, ActADD\mcite{turner2023steering} derives steering vectors as the difference of activations elicited by opposing prompts (e.g., truthful versus deceptive), while CAA\mcite{panickssery2023steering} averages activation differences between paired positive and negative prompts. To further improve steering effectiveness, another line of work\mcite{qiu2024spectral,zhang2024truthx,pham2024householder} explores performing interventions within learned, nonlinear subspaces. For instance, HPR\mcite{pham2024householder} proposes a direction-magnitude activation editing framework that learns global separating hyperplanes and rotations to reflect and rotate model activations toward desirable behaviors. 
However, most existing methods require externally annotated data (e.g., question-answer pairs). This reliance limits their adaptability to varying contexts and tethers their effectiveness to annotation quality.

{\bf Preference optimization.} Reinforcement learning from human feedback (RLHF) has emerged as a
prominent approach for learning human preferences\mcite{rlhf,lee2023rlaif}. RLHF first trains a reward model on preference data using established frameworks (e.g., the Bradley-Terry model\mcite{NIPS2004_825f9cd5}), and applies RL 
algorithms (e.g., PPO\mcite{ppo}) to optimize LLMs with respect to the reward model. Recent work\mcite{dpo,slic} shows the feasibility of bypassing the explicit reward modeling and directly solving the underlying RL problem. Further, SRSO\mcite{liu2024statistical} unifies the losses of DPO\mcite{dpo} and SLiC\mcite{slic}, offering an improved estimate of the optimal policy. This work extends previous research on preference optimization into challenging scenarios where externally annotated data is unavailable or impractical to obtain, addressing a critical gap in current work. 

{\bf LLM self-improvement.} 
Recent work\mcite{chen2025longpo, dong2024self, song2024mind, subramaniam2025multiagent, choi2024self, wu2024progress, peng2024regenesis, wan2025few} demonstrates the potential of enhancing LLM performance through self-improvement. By enabling models to generate, judge, and refine their own outputs, self-improvement has shown effectiveness across alignment, instruction-following, and preference modeling, often matching or surpassing fine-tuning while greatly reducing human annotation effort and exposure to harmful content. A spectrum of self-improvement methods have been proposed, including synthetic preference generation\mcite{dong2024self, lee2023rlaif}, tree-search refinement\mcite{cheng2024spar, lightstrategist}, Nash equilibrium-based optimization\mcite{wu2024self}, execution-guided verification\mcite{dong2024selfex}, and iterative self-evolved reward modeling\mcite{huang2024self}, which differ primarily in the mechanisms and granularity of self-generated feedback, ranging from internal judgment and strategic refinements to external execution validation. This work, to our best knowledge, represents the first exploration of this self-improvement paradigm in the context of model steering.

%% file: prelim.tex
\section{Preliminaries}
\label{sec:prelim}

\subsection{Model Steering}

Let $\mathcal{M}$ denote an $L$-layer, Transformer-based LLM and $x$ be a tokenized prompt.
The embedding matrix $W_E$ maps tokens to the initial hidden state $h_{0}=W_E(x)$.
For each layer $l\in [L]$, we apply multi-head attention (MHA) followed by a position-wise feed-forward network (FFN), each with a residual connection:\footnote{Layer normalization and projection matrices are omitted
for clarity.} 
\begin{equation}
h'_l = h_{l-1}
                   + \mathrm{MHA}_{l} \left (h_{l-1}\right),
\quad 
h_{l}        = h'_l
                   + \mathrm{FFN}_{l}\!\left(h^{\prime}_l\right).
\end{equation}
The model's logits are obtained via
$\mathcal{M}(x) =
      W_{U}(h_{L})$,
where $W_{U}$ is the un-embedding matrix.

During inference, we inject steering functions $f_{l}$ and $f'_{l}$ into the residual stream:
\begin{equation}
\tilde{h}'_l = \tilde{h}_{l-1}
      + \mathrm{MHA}_{l} (f_{l}(\tilde{h}_{l-1})),\quad 
\tilde{h}_l        = \tilde{h}'_l
      + \mathrm{FFN}_{l} (f'_{l}(\tilde{h}'_l)),
\end{equation}
where $f_{l}$ (respectively $f'_{l}$) operates immediately before the attention (respectively FFN) while the residual addition preserves the original signal. The steered model then produces $\tilde{\mathcal{M}}(x) =
      W_{U} (\tilde{h}_{L})$.

Given a dataset $\gD=\{(x_i,y^{+}_i,y^{-}_i)\}_{i=1}^{N}$,
where $y^{+}_i$ (desired) and $y^{-}_i$ (undesired) exhibit opposite attributes, we form positive and negative samples
$(x_i,y^{+}_i)$ and $(x_i,y^{-}_i)$, respectively. Passing these examples through $\mathcal{M}$ yields paired hidden activation sets:
\begin{equation}
\label{eq:pair}
\mathcal{H}_{l}^{+}=\{(h_{l,i}^{+},h_{l,i}^{\prime +})\}_{i},
\qquad
\mathcal{H}_{l}^{-}=\{(h_{l,i}^{-},h_{l,i}^{\prime -})\}_{i}.
\end{equation}
Existing model-steering methods learn $f_{l}$ and $f'_{l}$ by exploiting the discrepancy between $\gH_{l}^{+}$ and $\gH_{l}^{-}$ using contrastive or other representation-learning objectives (details in \msec{sec:relwk}). We refer to these methods as {\em steering-function learners} in the following.

\subsection{Self-Improvement Learning}

We formalize the self-improvement optimization as follow. Given an LLM $\gM$, we prompt $\gM$ with input $x$ and obtain two responses $y$ and $y'$. 

The self-improvement learning aims to optimize the alignment of $\mathcal{M}$ to human preferences. This process is typically done by reinforcement learning, which $\mathcal{M}$ represents the initial policy $\pi_0$. A preference oracle $\gO$, obtained from human feedback, is introduced in the learning process. Given the input $x$ and two responses $y$ and $y'$, The oracle $\gO$ will provide preference feedback $o(y \succ y' | x) \in \{0, 1\}$ indicating whether $y$ is preferred over $y'$. We denote $\mathbb{P}(y \succ y' | x) = \mathbb{E}\left[o(y \succ y' | x)\right]$ as the probability of $y$ `winning the duel' over $y'$. In addition, we define the winning probability of $y$ against a distribution of responses from policy $\pi$ as 
\begin{equation}
    \mathbb{P}(y \succ \pi | x) = \mathbb{E}_{y'\sim\pi(\cdot|x)}\left[\mathbb{P}(y' \succ y|x)\right].
\end{equation}
The self-improvement learning takes an iterative process to update the policy $\pi_t$, where $t$ denotes the iteration number. For every iteration $t$, $\pi_t$ is optimized based on the objective function as: 
\begin{equation}
\label{eq:update}
\pi_{t+1}
= \arg\min_{\pi}
\; \mathbb{E}_{y\sim\pi_t(\cdot| x)}
\Biggl[
\Bigl(
\log\frac{\pi(y| x)}{\pi_t(y| x)}
- \bigl(\eta\,\mathbb{P}(y\succ\pi_t| x) - \log Z_{\pi_t}(x)\bigr)
\Bigr)^2
\Biggr],
\end{equation}
where $\log Z_{\pi_t}(x)$ denotes the normalization term. Building upon this formulation, \sys extends the self-improvement paradigm to model steering, enabling LLMs to introspectively refine internal activations through iterative cycles of self-assessment and enhancement. 

%% file: method.tex
\section{Method}
\label{sec:method}

\begin{figure}[!t]
    \centering
    \includegraphics[width=\linewidth]{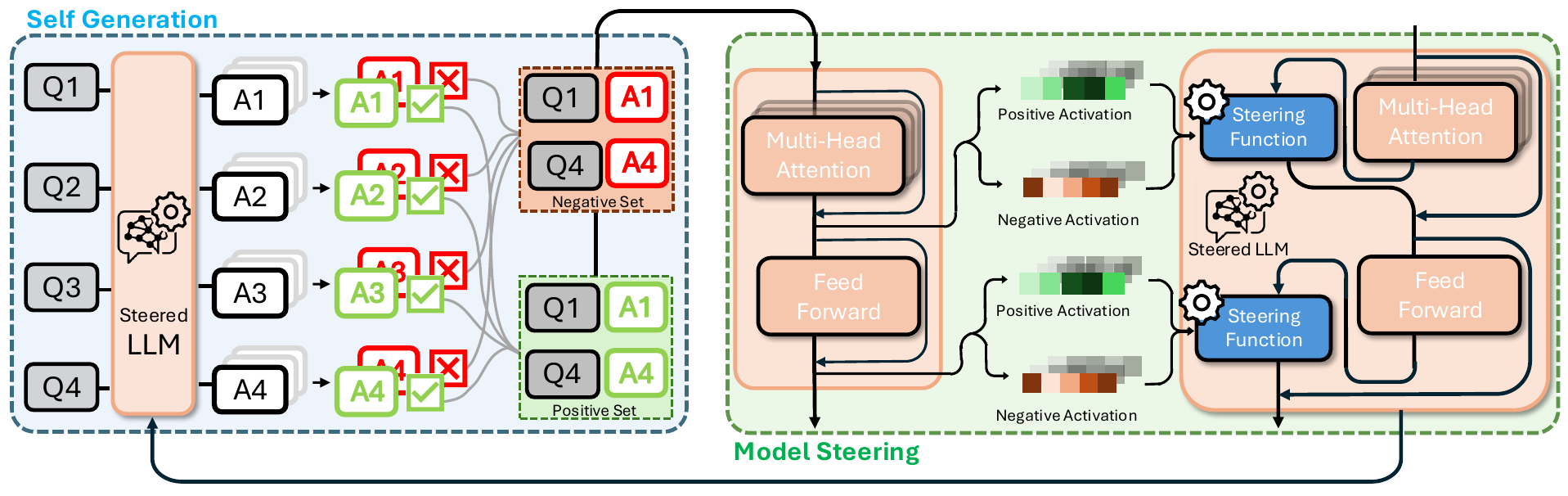}
    \caption{\small Overview of \sys. With $N = 4$ questions (prompt) drawn from a prompt distribution $\mathcal{D}_\text{prompt}$. We generate the $K = 3$ responses from model inference. We filter the responses with an oracle $\gO$ into a positive set and a negative set. Running these samples through the LLM, we collect the positive and negative activation sets. These sets are used to update the steering functions by the steering-function learner $\gA$. We combine the updated steering functions with the base model to form the refined policy for the next iteration. }
    \label{fig:pipeline}
\end{figure}

Next, we present \sys, the first self-improving model-steering framework, with its overview illustrated in Figure~\ref{fig:pipeline}.

\subsection{Self-Improving Model Steering}

At its core, \sys autonomously generates and refines contrastive samples through iterative self-improvement cycles, enabling learning the steering function from LLMs' own behaviors without external supervision.

At each iteration \(t\), the current steering policy \(\pi_{t-1}\) processes a mini-batch of \(N\) prompts sampled from the question distribution \(\gD_q\) (without requiring ground-truth answers that would necessitate human annotation).  For each prompt, the policy produces \(K\) candidate responses. A preference oracle $\gO$, which could be an existing reward model or even \(\pi_{t-1}\) itself acting as its own evaluator, is queried to yield an ordering over the \(K\) responses. These preference judgments define \emph{positive} (\(\mathcal{D}^{+}_{t}\)) and \emph{negative} (\(\mathcal{D}^{-}_{t}\)) sample buffers that pair each prompt with its preferred or disfavored outputs, respectively, creating contrastive training signals.

The language model \(\mathcal{M}\) is then executed on both positive samples
$(x_i,y^{+}_i)$ from $\gD^+_t$ and the negative samples $(x_i,y^{-}_i)$ from $\gD^-_t$. We collect layer-wise activations to construct two activation sets, \(\mathcal{H}^{+}_{l}\) and \(\mathcal{H}^{-}_{l}\), as defined in \meq{eq:pair}. We leverage an existing steering-function learner \(\mathcal{A}\) (e.g., HPR\mcite{pham2024householder}) to update the steering functions \(\{f_{l},\,f'_{l}\}_{l=1}^{L}\), which linearly or non-linearly shift model activations toward preferred behaviors while repelling undesirable ones. By composing the updated steering functions with the base model \(\mathcal{M}\), we derive the refined policy \(\pi_{t}\) for the next iteration, effectively solving the optimization problem in \meq{eq:update}.

The above process is iteratively repeated to progressively refine the steering functions. Because \sys bootstraps its training signal entirely from its own generated outputs, it decouples model steering from externally annotated data and can be extended through an arbitrary number of iterations \(T\). Under mild assumptions about oracle accuracy, the policy sequence \(\{\pi_{t}\}_{t=0}^{T}\) constitutes monotonic improvement in expected preference reward. Crucially, each update operates only on sub-token activations rather than modifying full model weights, thereby maintaining computational efficiency compared to full-scale fine-tuning.

The complete algorithm is sketched in Algorithm~\ref{alg:sims1}.

\begin{algorithm}\small
\caption{\textsc{Self-Improving Model Steering (SIMS)}}
\label{alg:sims1}
\KwIn{Language model $\mathcal{M}$ with $L$ layers;   
           preference oracle $o$; steering-rule learner $\mathcal{A}$;   
           prompt distribution $D_{\!\text{prompt}}$; iterations $T$;   
           prompts per iteration $N$; responses per prompt $K$}
\KwOut{Final steered policy $\pi_{T}$}
Initialize steering transforms
 $\{f^{(0)}_{l}\}_{l=1}^{L}$ and $\{f'^{(0)}_{l}\}_{l=1}^{L}$\;
Define initial policy $\pi_{0}=(\mathcal{M},\{f^{(0)}_{l}\}_{l=1}^{L},\{f'^{(0)}_{l}\}_{l=1}^{L})$\;
  \For{$t=1$ \textbf{to} $T$}{
  Sample prompts $\{\mathbf{x}_{n}\}_{n=1}^{N}\!\sim\!D_{\!\text{prompt}}$\;
    \For{$n=1$ \textbf{to} $N$}{
         Generate $K$ candidate responses
        $\{\mathbf{y}_{n,k}\}_{k=1}^{K}\!\sim\!\pi_{t-1}(\,\cdot\mid\mathbf{x}_{n})$\;
        Query oracle $o$ for pairwise preferences
  $\mathbb{P}_o(\mathbf{y}_{n,k}\succ\mathbf{y}_{n,k'})$
  \tcp*{All $k\!<\!k'$}
        }
Construct datasets
$\mathcal{D}^{+}_{t}=\bigl\{(\mathbf{x}_{n},\mathbf{y}_{n,k})\mid
              \mathbb{P}(y_{n,k} \succ \bm{\pi}_t \mid x_{n}) \bigr\},
        \quad
        \mathcal{D}^{-}_{t}=\bigl\{(\mathbf{x}_{n},\mathbf{y}_{n,k})\mid
               \mathbb{P}(y_{n,k} \prec \bm{\pi}_t \mid x_{n}) \bigr\}$\;
 Collect hidden activations
 $\mathcal{H}^{+}_{l}=\{\mathcal{M}_{l}(\mathbf{x},\mathbf{y})\}_{(\mathbf{x},\mathbf{y})\in\mathcal{D}^{+}_{t}},\;
           \mathcal{H}^{-}_{l}=\{\mathcal{M}_{l}(\mathbf{x},\mathbf{y})\}_{(\mathbf{x},\mathbf{y})\in\mathcal{D}^{-}_{t}}$\;
 Learn new steering functions
$\bigl\{f^{(t)}_{l},f'^{(t)}_{l}\bigr\}_{l=1}^{L}
           =\mathcal{A}\bigl(\mathcal{H}^{+}_{1:L},\mathcal{H}^{-}_{1:L}\bigr)$\;
Update policy
     $\pi_{t}=(\mathcal{M},\{f^{(t)}_{l}\}_{l=1}^{L},\{f'^{(t)}_{l}\}_{l=1}^{L})$\;
  }
  \Return $\pi_{T}$\;
\end{algorithm}

\subsection{Prompt Ranking and Contrast Sampling}

To further improve \sys's adaptability and effectiveness, we introduce two optimization strategies. 

To eliminate dependency on external reward models, we introduce {\em prompt ranking} (\syspr), a fully self-supervised alternative that leverages the model's own judgment capabilities to generate preference signals. For each prompt $x_{i}$ at the $t$-th iteration, we query the current policy $\pi_{t-1}$ for $K$ candidate completions $\{y_{i,k}\}_{k=1}^{K}$ as in the original iteration loop. Instead of passing pairs to the oracle, we instruct the backbone model $\mathcal{M}$ to \emph{rank} the complete set of responses under an instruction (ranking) prompt. The highest-ranked responses form the positive set $\mathcal{D}^{+}_{t}$, while the lowest-ranked ones populate the negative set $\mathcal{D}^{-}_{t}$. These contrastive samples are fed to the steering-function learner $\mathcal{A}$ following the same protocol as the standard \sys. The implementation details are deferred to \msec{sec:app}.

Orthogonally, to further improve the sample quality for steering-function learning, we introduce {\em contrast sampling} (\syscs), a strategy that reuses previous responses but selects only the most informative question-response pairs for the steering-function learner. 

Specifically, for each prompt $x_i$, we compute a margin-style reward:
\begin{equation}
  r_i \;=\;
     \underbrace{\max_{k}\,\mathbb{P}_o\bigl(y_{i,k}\succ\pi_{t}| x_i\bigr)}
     _{\text{best candidate}}
     \;-\;
     \underbrace{\max_{k}\,\mathbb{P}_o\bigl(y_{i,k}\prec\pi_{t}| x_i\bigr)}
     _{\text{worst candidate}} ,
\end{equation}
which rewards the most positive completion and penalizes the most negative one. After scoring each prompt $x_i$, the triple $(x_i,\{y_{i,k}\}_{k=1}^{K},r_i)$ is appended to a memory bank $\mathcal{B}$, which stores the prompts and responses from the previous iterations. This replay-like procedure helps \sys to better utilize the preference signals from the oracle.

At the beginning of each steering update, we sample \(           \mathcal{D}_t = \mathrm{top}_N(\mathcal{B})         \), the $N$ tuples in $\mathcal{B}$ with the highest contrast reward. For each retained prompt, the highest-ranked completion forms a positive pair and the lowest-ranked completion forms a hard negative pair:
\begin{equation}
  \mathcal{D}_t^{+}=\bigl\{(x_i,y_{i,(1)})\bigr\}, \qquad
  \mathcal{D}_t^{-}=\bigl\{(x_i,y_{i,(K)})\bigr\},
\end{equation}
The following steps are the same as the standard \sys to update the steering functions. The implementation details of \syscs are deferred to \msec{sec:app}.

%% file: exps.tex
\section{Evaluation}
\label{sec:eval}

\subsection{Experimental Setting}

\textbf{Datasets.} 
We employ the UltraFeedback corpus\mcite{cui2023ultrafeedback} as the primary evaluation dataset. UltraFeedback consists of 64\,000 prompts, each paired with multiple candidate responses ranked by human annotators. For conventional model-steering methods that require supervised preference data, we use the complete prompt-response pairs with their associated rankings.
When evaluating \sys, we deliberately discard all responses and rankings, using only the raw prompts.

\textbf{Models.} We evaluate \sys on two representative LLMs: {\tt llama3-8b}\mcite{grattafiori2024llama} and {\tt mistral-7b}\mcite{jiang2023mistral7b}. Due to computational limitations, if otherwise specified, we limit the model's response to $128$ tokens. We employ PairRM\mcite{llm-blender-2023} as the oracle in our experiments. We use HPR\mcite{pham2024householder} as the underlying steering-function learner for both \sys and baselines.

\textbf{Metrics.} We use Alpaca-Eval\mcite{dubois2025lengthcontrolledalpacaevalsimpleway} and Arena-Hard\mcite{li2024crowdsourced} to evaluate the performance of post-steering models in open-ended question answering. For Alpace-Eval, we report two complementary metrics: WinRate (WR) and length-control WinRate (LC). WR is defined as the average preference probability of a given model over {\tt gpt-4-turbo}, as judged by {\tt gpt-4o}\mcite{openai2024gpt4ocard}. LC refines WR by applying a causal logistic-regression adjustment to neutralize answer-length biases, yielding counterfactual, equal-length win probabilities. For Arena-Hard, we implement the following comparison protocol:
comparing the model's outputs and {\tt gpt-3.5-turbo}'s answers on 500 challenging prompts (each judged twice with position swapping), mapping {\tt gpt-4o}'s 5-point Likert preferences to wins/losses, fitting a Bradley–Terry model to these 1,000 pairwise results, and reporting the bootstrap-estimated win-rate (with confidence interval) against the baseline.

\subsection{Main Results}

\begin{figure}[t]
    \centering
    \includegraphics[width=0.95\linewidth]{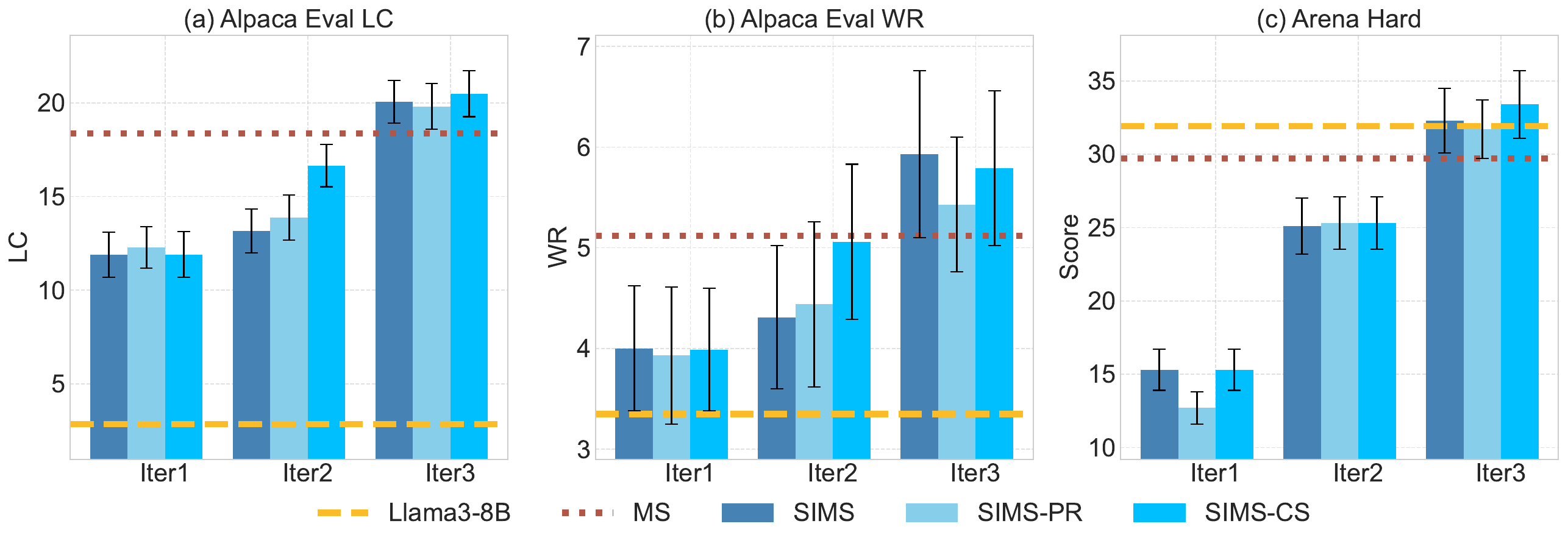}
    \caption{\small {\tt llama3-8b} under model steering versus three iterations of \sys, optionally enhanced with \syspr or \syscs.  Reported are length-controlled win-rate (LC), win-rate (WR), and Arena-Hard score (higher is better; mean $\pm$ s.d.).  \syscs on Iter 3 attains the strongest overall performance.}
    \label{fig:base_l}
\end{figure}

\begin{figure}[!t]
    \centering
    \includegraphics[width=0.95\linewidth]{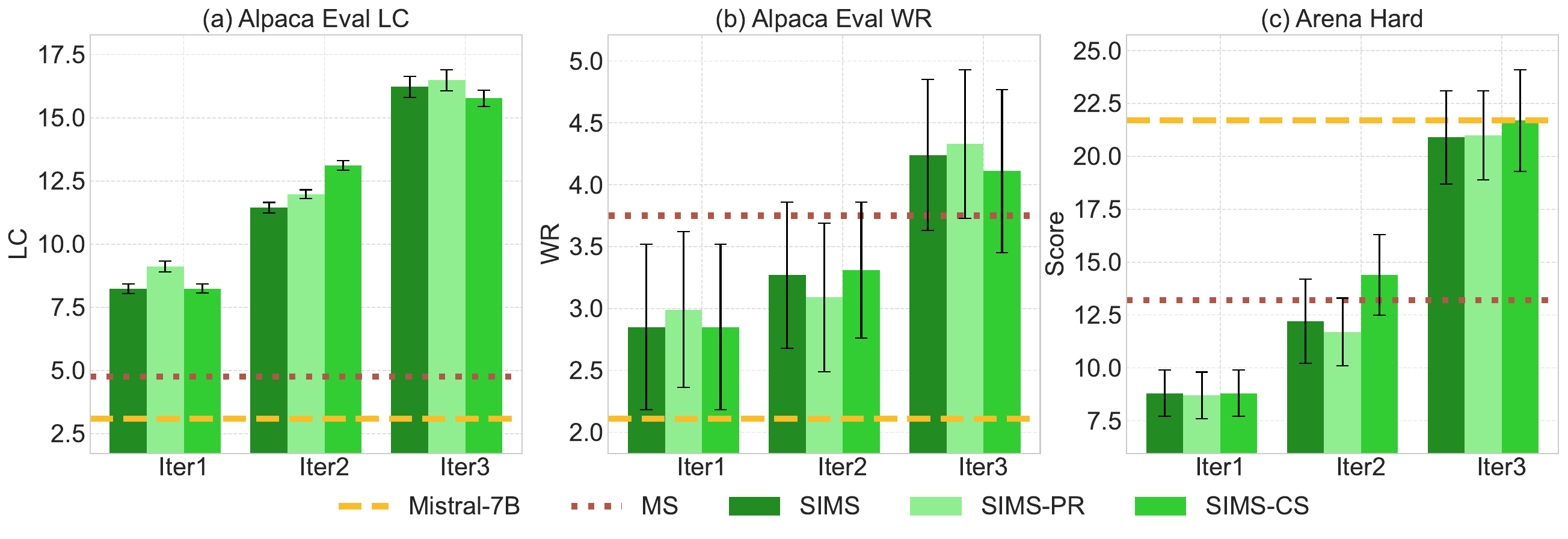}
    \caption{\small {\tt mistral-7b} under model steering versus three iterations of \sys, optionally enhanced with \syspr or \syscs.  Reported are length-controlled win-rate (LC), win-rate (WR), and Arena-Hard score (higher is better; mean $\pm$ s.d.).  \syscs on Iter 3 attains the strongest overall performance.}
    \label{fig:base_m}
\end{figure}

Figure~\ref{fig:base_l} presents the results on {\tt llama3-8b}. Notably, at Iter1, \sys elevates the LC WinRate of the base model ({\tt llama3-8b}) from 2.86 to 11.89 (315\% increase). Similarly, the WR score rises from 3.35 to 4.00. Further, at Iter2, \sys observes consistent and significant growth across all metrics. Its LC WinRate increases to 13.16 (+10.7\% over Iter1), its WR improves to 4.31 (+7.8\%), and its Arena-Hard performance surges to 25.1 (+64\%). The enhanced variant \syscs, in particular, shows significant improvement with its LC WinRate jumping to 16.65 and WR reaching 5.06, suggesting that the contrastive sampling strategy successfully identifies more informative samples to accelerate representation refinement. Finally, at Iter3, \sys outperforms conventional model steering that relies on annotated data, by 1.70 on LC, 0.81 on WR, and 2.6 on Arena-Hard. The peak performance appears among the variants of \sys: \syscs achieves 20.49 on LC and 33.4 on Arena-Hard, while \sys reaches 5.79 on WR, validating our core hypothesis about the viability and advantages of self-improving model steering. 
Sample outputs of different steered models are deferred to \msec{sec:samples}.

Figure~\ref{fig:base_m} illustrates the experimental results on {\tt mistral-7b}, which closely parallel the findings from the evaluation on {\tt llama3-8b}. Consistent with our previous observations, \sys demonstrates robust performance gains across all metrics (WR, LC, and Arena-Hard), exhibiting steady improvement trajectories through successive iterations. 

\begin{figure}
    \centering
    \includegraphics[width=\linewidth]{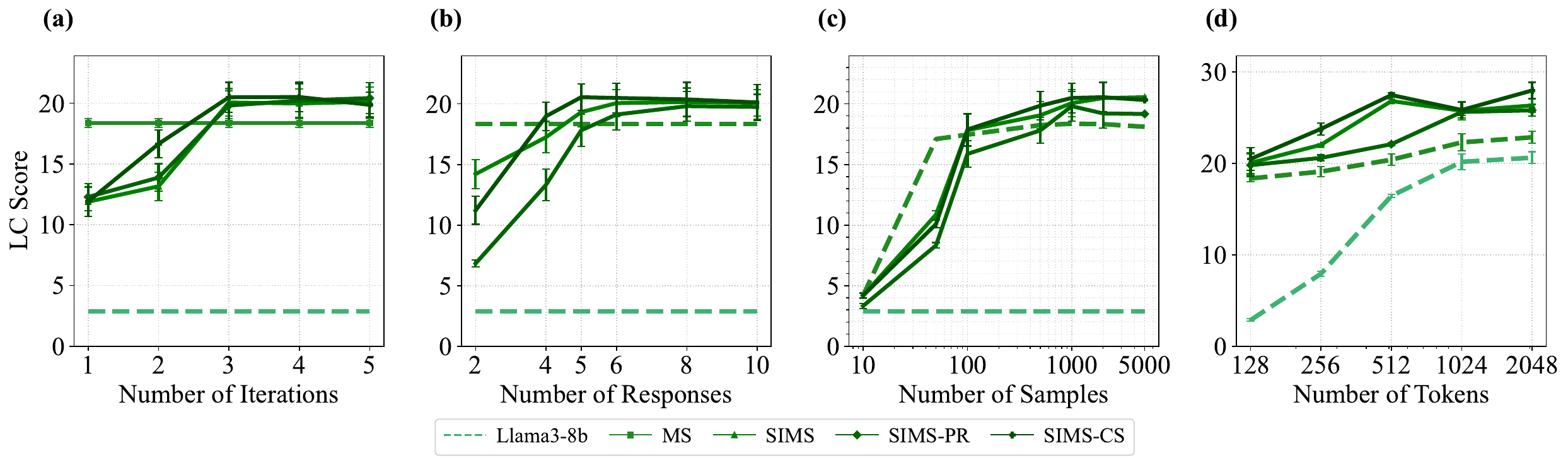}
    \caption{\small LC results of (a) number of samples, (b) number of responses, (c) number of samples, and (d) number of tokens based on {\tt llama3-8b}.}
    \label{fig:lc_abl}
\end{figure}

\begin{figure}
    \centering
    \includegraphics[width=\linewidth]{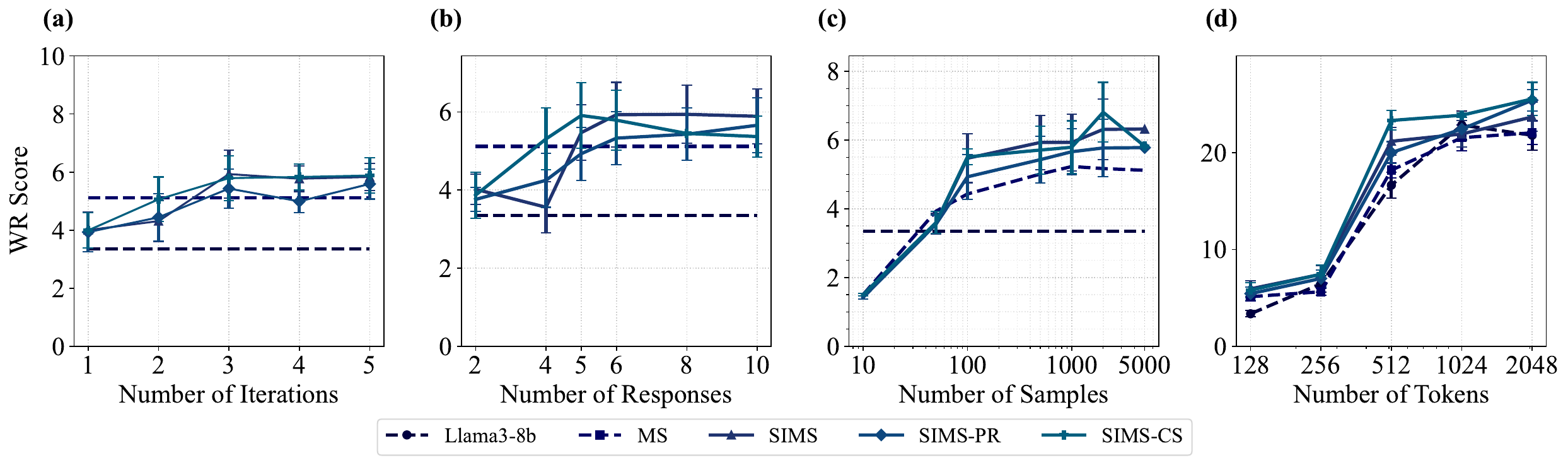}
    \caption{\small WR results of (a) number of samples, (b) number of responses, (c) number of samples, and (d) number of tokens based on {\tt llama3-8b}.}
    \label{fig:wr_abl}
\end{figure}

\subsection{Ablation Study}

We further conduct an ablation study to explore how different factors impact \sys's performance.

Our enhanced variants exhibit even more efficient optimization trajectories. At iteration 1, \syspr achieves scores of 12.28/3.93 while \syscs reaches 11.91/3.99. Notably, \syscs reaches peak performance earlier (Iter 4: 20.51/5.83), slightly outperforming \syspr's best results (Iter 5: 20.42/5.59). Both variants demonstrate performance plateaus or slight declines thereafter, with \syscs dropping marginally to 19.87/5.88 at iteration 5.
These convergence patterns suggest that continuing iterations beyond the point where steering has stabilized provides diminishing returns. Based on this comprehensive analysis, we recommend an optimal configuration of three iterations for most applications, balancing performance gains with computational efficiency.

{\bf \# Iterations.} Figure~\ref{fig:lc_abl}\,(a) and Figure~\ref{fig:wr_abl}\,(a) illustrate the impact of varying the number of iterations. While the baseline {\tt llama-3-8b} and the conventional model steering maintain fixed LC/WR scores at 2.86/3.35 and 18.36/5.12, respectively (as non-iterative approaches), \sys demonstrates significant performance improvements across successive iterations. From Iter1 to Iter3, LC increases from 11.89 to 20.06 and WR from 4.00 to 5.93. Its performance then stabilizes in subsequent iterations (Iter4: 19.98/5.77, Iter5: 20.12/5.83), indicating that most performance benefits are realized within the first four rounds of self-improvement. The enhanced variants \syspr and \syscs exhibit even more efficient optimization trajectories. At Iter1, \syspr attains 12.28/3.93, while \syscs reaches 11.91/3.99. Notably, \syscs reaches peak performance earlier (Iter 4: 20.51/5.83), slightly outperforming \syspr's best result (Iter5: 20.42/5.59). The performance of both variants plateaus or declines slightly thereafter.
These convergence patterns suggest that continuing iterations beyond the point where steering has stabilized provides diminishing returns.

\textbf{\# Responses.} Figure~\ref{fig:lc_abl}\,(b) and Figure~\ref{fig:wr_abl}\,(b) illustrate how \sys's performance scales with the number of candidate responses generated during the steering process. Varying the number of sampled responses $K$ reveals a clear trade-off between sample size and alignment quality. With the standard \sys, performance increases consistently from 14.21/4.02 (LC/WR) at $K$ = 2 to 20.16/5.94 at $K$ = 10, demonstrating that more diverse candidates enable more precise alignment through richer contrastive signals. The enhanced variants \syspr and \syscs further amplify the performance gains. \syscs yields the larger early boost, at $K$ = 2, it outperforms \syspr (11.21/3.87 versus 6.83/3.76) and retains the highest LC throughout, ultimately reaching 20.12/5.37 at $K$ = 10.  Interestingly, \syspr demonstrates slightly superior WR performance at the highest sampling rate (19.75/5.66 at $K$ = 10). The optimal performance is achieved by \syscs with $K$ = 10, substantially outperforming both the baseline {\tt llama-3-8b} (17.86/2.02) and conventional model steering (1.76/2.50). These results confirm that \mct{i} \sys's effectiveness scales with the diversity of candidate responses, and \mct{ii} \syscs is the most effective method, especially when the response budget is small.

\textbf{\# Samples.} We vary the sample size from the prompt dataset from 10 to 5,000. Figure~\ref{fig:lc_abl}\,(c) and Figure~\ref{fig:wr_abl}\,(c) reveal a clear performance hierarchy. \sys shows strong scaling properties, exhibiting nearly monotonic improvement: from 4.18/1.50 (LC/WR) with 10 samples, 11.02/3.94 with 1,000 samples, and 20.55/6.32 with 5,000 samples. This consistent scaling confirms that the iterative, self-generated feedback effectively leverages additional data to refine the steering vectors. In contrast, conventional model steering shows minimal sensitivity to corpus size, marginally improving from 18.36 to 19.12 on LC and from 5.12 to 5.24 on WR. This saturation pattern indicates that fixed steering approaches quickly exhaust learning capacity without iterative refinement.

\syspr and \syscs further amplify the self-improving gains, with \syscs consistently leading other approaches. Even in an extremely data-constrained setting (10 samples), \syscs outperforms \syspr (4.18/1.50 versus 3.32/1.42). The gap widens with more samples: at 2,000 samples, \syscs attains 20.55/6.81 compared to \syspr's 19.21/5.78, maintaining its lead across all sizes. This can be attributed to \syscs's ability to harvest broader, more diverse response pools, enabling more effective selection of high-quality contrastive examples. Overall, while all \sys variants benefit from larger datasets, \syscs emerges as the most effective approach for leveraging additional data diversity to enhance model steering.

\textbf{Token length.} We further analyze the impact of response token length. Figure~\ref{fig:lc_abl}\,(d) and Figure~\ref{fig:wr_abl}\,(d) reveal a clear length-dependent performance pattern. For the baseline {\tt llama3-8b}, both LC and WR increase steadily from 2.86/3.35 at 128 tokens to 20.63/21.80 at 2,048 tokens, confirming that longer contexts lead to higher-quality responses. Conventional model steering shifts the performance curve upward (18.36/5.12 at 128 tokens and 22.86/22.05 at 2,048 tokens), showing that steering advantages are potentially amplified with increasing context length. The variants of \sys yield the most substantial performance enhancements across all context lengths. Standard \sys achieves 20.06/5.93 (128) and 26.35/23.69 (2048); \syspr provides additional improvement (e.g., 26.91/24.42 at 2,048 tokens), while \syscs consistently leads across all context lengths, peaking at 27.99/25.55 for the full-length setting. Overall, the performance of all methods scales with context length, while \syscs emerges as the most effective method for leveraging increased context.

\begin{wrapfigure}{r}{0.5\textwidth}
    \begin{center}
        \includegraphics[width=0.5\textwidth]{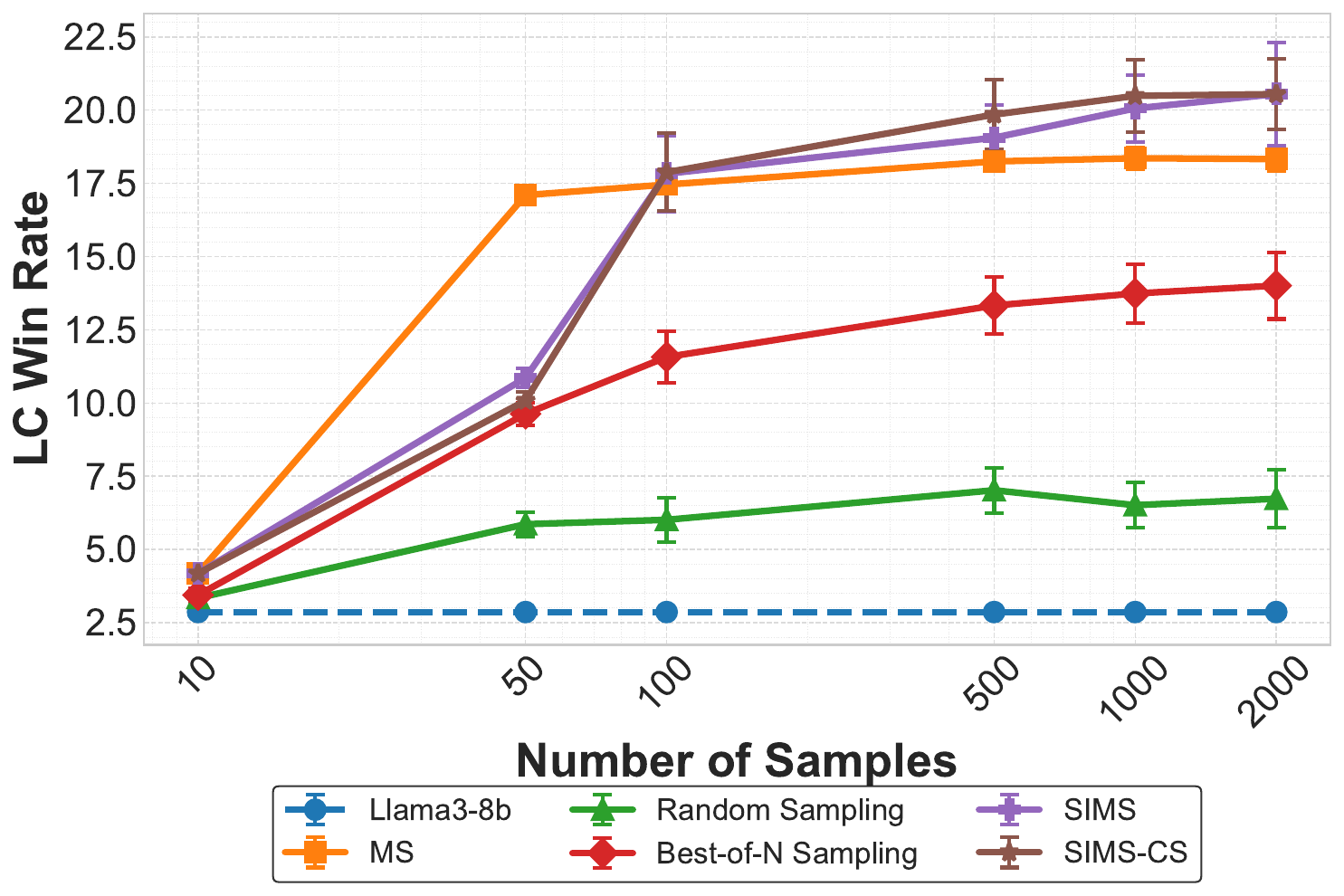}
    \end{center}
    \caption{\small Impact of sampling strategy. }
    \label{figs:sample2win}
\end{wrapfigure}

\subsection{Exploration}

\textbf{Sampling strategy.} We show the influence of sampling strategies on steering performance in Figure~\ref{figs:sample2win}. We compare \sys (oracle-based) and \syscs (contrast sampling) with two naive sampling strategies, random sampling and best-of-$N$ sampling. For random sampling, responses for each sample are selected randomly as positive or negative. Although random sampling doubles LC to 6.03 with 500 samples, it quickly saturates, indicating that unguided data accumulation provides limited steering signals. For best-of-$N$ sampling, we collect 10 random samples and pick the one with the highest LC. Best-of-$N$ outperforms random sampling (6.26 with 500 samples, 11.69 with 2,000 samples). The improvement saturates after 500 samples, suggesting that best-of-$N$ captures only coarse preference improvements. In contrast, \sys rises steadily to 20.06, while \syscs leverages contrastive sampling to edge higher, reaching 20.49 with 1,000 samples.

\begin{wrapfigure}{r}{0.5\textwidth}
    \begin{center}
    \includegraphics[width=0.5\textwidth]{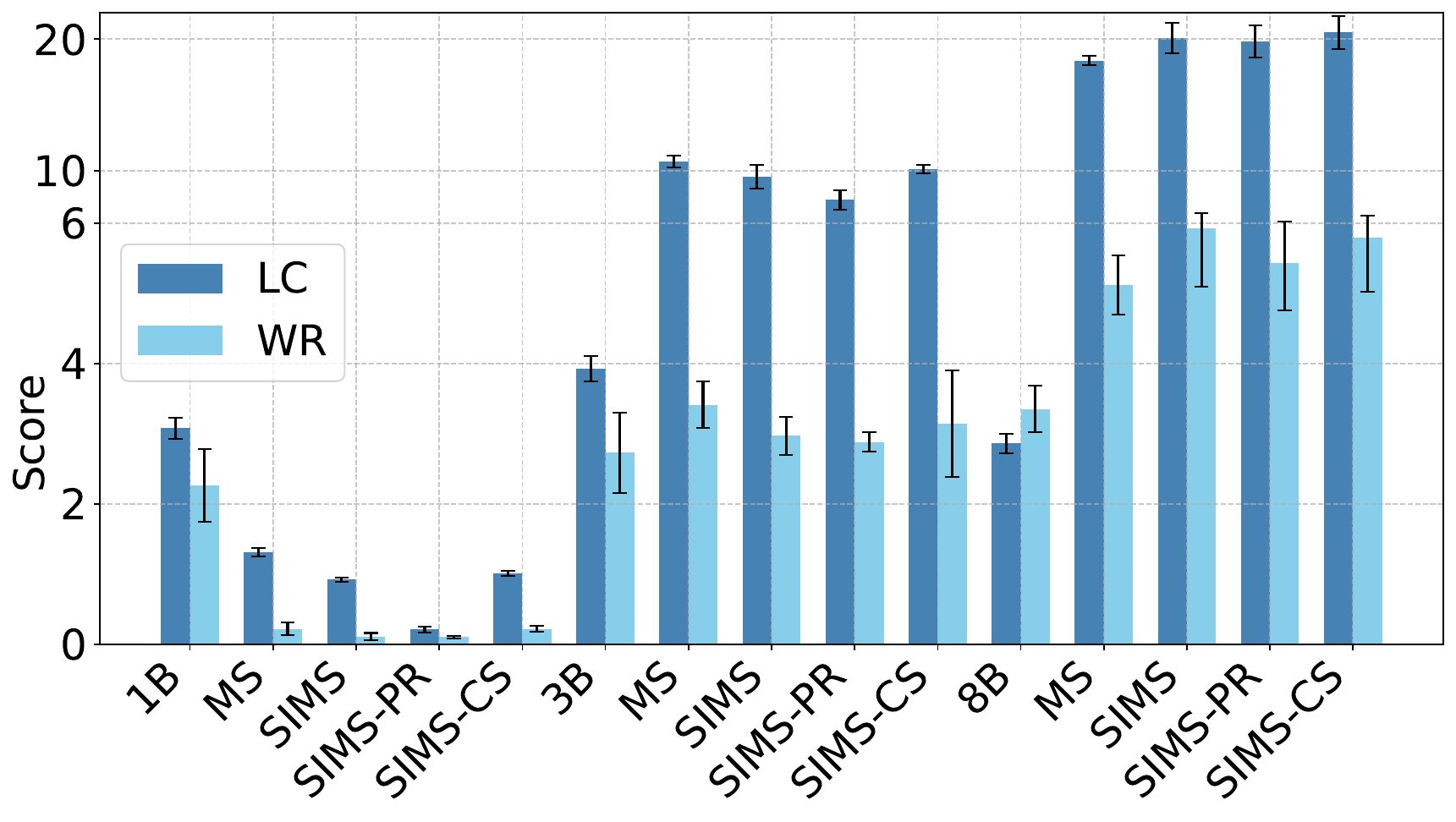}
    \end{center}
    \caption{\small Impact of LLM size varying from 1B to 8B.}
    \label{fig:model}
    \vspace{-5pt}
\end{wrapfigure}

\textbf{LLM scale.} Figure~\ref{fig:model} illustrates how \sys performance scales as the backbone LLM size increases from 1B to 8B, revealing a strong correlation between steering effectiveness and the underlying model's capabilities. With the smallest 1B model, all three steering variants show marginal effectiveness, achieving only minimal scores (LC = 0.92, WR = 0.11). This performance limitation stems from the model's inherent constraints: it typically generates brief, repetitive continuations that provide insufficient variation for the steering learner to extract robust and stable directional vectors. In comparison, the 8B model generates substantially longer, more coherent responses with a 
wider quality distribution, revealing clearer and more informative preference signals. Under identical configurations, all variants achieve higher performance (20.06 LC, 5.93 WR). Although the relative improvement from 3B to 8B appears less dramatic than the transition from 1B to 3B, the absolute performance gains remain substantial. This scaling pattern shows that self-generated steering continues to benefit from increased model scale: once the model is capable of producing sufficiently nuanced and diverse outputs, the learning algorithm can effectively distill stronger and more precise steering.

\textbf{Alternative tasks.} Beyond open-ended question answering, we further validate \sys's generalizability on 8 NLP benchmarks spanning a range of capabilities: deductive and commonsense reasoning (ARC\mcite{allenai:arc}, Winogrande\mcite{sakaguchi2021winogrande}, and HellaSwag\mcite{zellers2019hellaswag}); open-domain question answering (TriviaQA\mcite{joshi2017triviaqa}); broad knowledge transfer (MMLU\mcite{hendrycks2020measuring}); sentiment analysis (SST-2\mcite{socher-etal-2013-recursive}); and safety \& security (TruthfulQA\mcite{lin2021truthfulqa} and ToxiGen\mcite{hartvigsen2022toxigen}). We randomly draw prompts from the available training pool at every iteration. Because MMLU and TruthfulQA lack official training splits, we divide each benchmark’s public items into two non-overlapping subsets of equal size, using only the first subset for training and reserving the second for evaluation.

\begin{wrapfigure}{r}{0.5\textwidth}
    \begin{center}
    \vspace{-5pt}
    \includegraphics[width=0.5\textwidth]{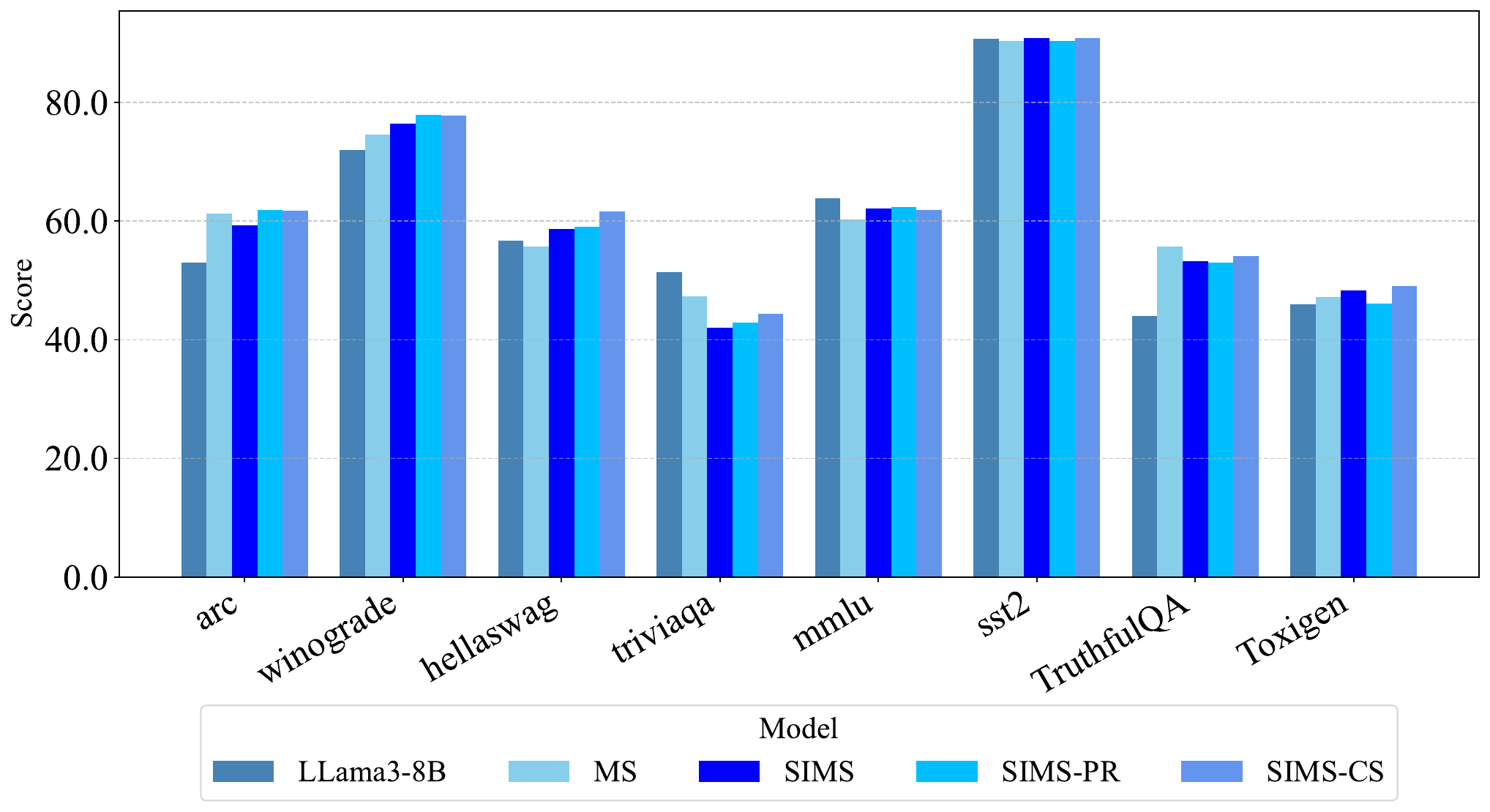}
    \end{center}
    \caption{\small Performance of \sys on 8 NLP benchmarks. }
    \label{fig:tasks}
\end{wrapfigure}
Starting from {\tt llama3-8b}, baseline model steering raises the average score by +1.9. However, its gains manifest unevenly across different task categories: while reasoning-focused tasks such as ARC (+8.3) and Winogrande (+2.6) show substantial improvement, knowledge-intensive tasks such as HellaSwag (–0.9) and TriviaQA (–4.1) regress. This inconsistency suggests that a conventional steering vector cannot accommodate disparate task requirements. Our self-improving method elevates the average to 62.0 without external labels by iteratively exploring the model's intrinsic representation space. 

The enhanced variants further amplify these gains: \syspr guides the learner toward more informative preference gradients, raising average performance to 62.5, while \syscs enhances learning by supplying more challenging negative examples that expand the coverage of steering directions, achieving the highest overall score of 63.6, an improvement of $3.9$ over the base model.

%% file: conclude.tex
\section{Conclusion and Future Work}
\label{sec:conclude}

This paper presents \sys, the first self-improving model-steering framework that operates without external supervision. At its core, \sys autonomously generates and evaluates contrastive samples through iterative self-improvement cycles, enabling adaptive, context-specific steering. Extensive empirical evaluation demonstrates \sys's effectiveness, consistently outperforming or matching state-of-the-art steering methods that rely on external annotations.

While this work highlights self-improving model steering as a promising direction for future research on inference-time LLM alignment, several limitations warrant further investigation. First, we only evaluate \sys on the language-based tasks. A further analysis on other modalities (e.g., vision) is needed to validate \sys's generalization. Second, we evaluate \sys method based on existing steering-function learners. Future work could explore learners specifically optimized for the self-improving steering framework. Third, future work could also improve the prompt ranking and contrast sampling strategies. For instance, one could apply in-context learning when ranking prompts, which provides supportive information for LLMs to better evaluate self-generated responses, leading to higher-quality samples for learning steering functions.

%% file: app.tex
\appendix

\section{Implementation Details}
\label{sec:app}


\begin{algorithm}\small
\caption{\textsc{SIMS} with Prompt Ranking (\textsc{SIMS–PR})}
\label{alg:sims}
\KwIn{Language model $\mathcal{M}$ with $L$ layers;   
           preference oracle $o$; steering-rule learner $\mathcal{A}$;   
           prompt distribution $D_{\!\text{prompt}}$; iterations $T$;   
           prompts per iteration $N$; responses per prompt $K$, ranking prompt $\bm{p}$}
Initialize steering transforms
        $\{f^{(0)}_{l}\}_{l=1}^{L}$ and $\{f'^{(0)}_{l}\}_{l=1}^{L}$\;
Define initial policy
        $\pi_{0}=(\mathcal{M},\{f^{(0)}_{l}\}_{l=1}^{L},\{f'^{(0)}_{l}\}_{l=1}^{L})$\;
\For{$t=1$ \textbf{to} $T$}{
    Sample prompts $\{\mathbf{x}_{n}\}_{n=1}^{N}\!\sim\!D_{\!\text{prompt}}$\;
      \For{$n=1$ \textbf{to} $N$}{
        Generate $K$ candidate responses
                 $\{\mathbf{y}_{n,k}\}_{k=1}^{K}\!\sim\!\pi_{t-1}(\,\cdot\mid\mathbf{x}_{n})$\;
        \textbf{Prompt Ranking}: Query $\mathcal{M}$ to rank all $K$ response based on the prompt $\mathbf{x}$ and ranking prompt $\bm{p}$ as  
$\mathcal{M}\!\left(
      \mathbf{y}_{n,1},\dots,\mathbf{y}_{n,K}
      \;\middle|\;
      \mathbf{x}_{n},\bm{p}
  \right)
  \;\longrightarrow\;
  \mathbf{y}_{n,(1)} \succ \mathbf{y}_{n,(2)} \succ \cdots \succ \mathbf{y}_{n,(K)}$\;
}
 Construct datasets
        $\mathcal{D}^{+}_{t}=\bigl\{(\mathbf{x}_{n},\mathbf{y}_{n,(1)})\},
          \quad
          \mathcal{D}^{-}_{t}=\bigl\{(\mathbf{x}_{n},\mathbf{y}_{n,(K)})\bigr\}$\;
    Collect hidden activations
        $\mathcal{H}^{+}_{l}=\{\mathcal{M}_{l}(\mathbf{x},\mathbf{y})\}_{(\mathbf{x},\mathbf{y})\in\mathcal{D}^{+}_{t}},\;
          \mathcal{H}^{-}_{l}=\{\mathcal{M}_{l}(\mathbf{x},\mathbf{y})\}_{(\mathbf{x},\mathbf{y})\in\mathcal{D}^{-}_{t}}$\;
    Learn new steering functions
          $\bigl\{f^{(t)}_{l},f'^{(t)}_{l}\bigr\}_{l=1}^{L}
          =\mathcal{A}\bigl(\mathcal{H}^{+}_{1:L},\mathcal{H}^{-}_{1:L}\bigr)$\;
      Update policy
        $\pi_{t}=(\mathcal{M},\{f^{(t)}_{l}\}_{l=1}^{L},\{f'^{(t)}_{l}\}_{l=1}^{L})$\;
}
  \Return $\pi_{T}$
\end{algorithm}

The goal of \syspr is to iteratively steer a pretrained language model $\mathcal{M}$ toward a desired behaviour without any external supervision. It achieves this by replacing the human or task-specific preference oracle from the original \sys algorithm with a ranking prompt that the model executes on its outputs. This change yields an oracle-free preference signal, enable a more efficient self-improving model steering.

Let $\pi_{t}=(\mathcal{M},\{f^{(t)}_{l}\}_{l=1}^{L},\{f'^{(t)}_{l}\}_{l=1}^{L})$ denote the steered policy at iteration $t$, where $f^{(t)}_{l},f'^{(t)}_{l}\!:\!\mathbb{R}^{d}\!\to\!\mathbb{R}^{d}$ are layer-wise activation transforms learnt so far. At every step we draw $N$ prompts $\mathbf{x}_{1:N}\!\sim\!D_{\!\text{prompt}}$ and elicit $K$ candidate continuations $\mathbf{y}_{n,1:K}\!\sim\!\pi_{t-1}(\cdot\mid\mathbf{x}_{n})$. Rather than querying an external oracle for comparisons, we issue a ranking call to the backbone model:
$$  
\mathcal{M}\!\left(
      \mathbf{y}_{n,1},\dots,\mathbf{y}_{n,K}
      \;\middle|\;
      \mathbf{x}_{n},\bm{p}
  \right)
  \;\longrightarrow\;
  \mathbf{y}_{n,(1)} \succ \mathbf{y}_{n,(2)} \succ \cdots \succ \mathbf{y}_{n,(K)},
$$
where $\bm{p}$ is a \emph{task-agnostic ranking prompt} (refer to the following as an example). The call returns a ranking over the $K$ candidates.
We then keep
\[
  \mathcal{D}^{+}_{t}
    = \!\Bigl\{\!\bigl(\mathbf{x}_{n},\mathbf{y}_{n,(1)}\bigr)
              \Bigr\},\quad
  \mathcal{D}^{-}_{t}
    = \!\Bigl\{\!\bigl(\mathbf{x}_{n},\mathbf{y}_{n,(K)} \Bigr\},
\]
The sets play the same role as oracle-labelled \emph{wins} and \emph{losses} in \sys, but do not need additional oracle model and improve the efficiency.

For every layer $l$, we collect hidden activations
\[
  \mathcal{H}^{+}_{l}
    =\Bigl\{\mathcal{M}_{l}(\mathbf{x},\mathbf{y})\Bigr\}_{(\mathbf{x},\mathbf{y})\in\mathcal{D}^{+}_{t}},\qquad
  \mathcal{H}^{-}_{l}
    =\Bigl\{\mathcal{M}_{l}(\mathbf{x},\mathbf{y})\Bigr\}_{(\mathbf{x},\mathbf{y})\in\mathcal{D}^{-}_{t}},
\]
and invoke the steering learner
\(
  \bigl\{f^{(t)}_{l},f'^{(t)}_{l}\bigr\}_{l=1}^{L}
  =\mathcal{A}\bigl(\mathcal{H}^{+}_{1:L},\mathcal{H}^{-}_{1:L}\bigr).
\)
This step is identical to \sys.

\begin{mtbox}{}
    \small
        \textbf{Prompt}: I want you to create a leaderboard of large-language model's responses. To do so, I will give you the instructions (prompts) given to the model, and the responses of model. To make a leaderboard, first make a list ranking which responses would be preferred by humans, then give the resulting list of JSON to `make leaderboard`.
        Here is the prompt: \\
        \{\{ \\
            "instruction": "{instruction}", \\
        \}\} \\
        Here is the responses from the model:  [ \\ \{response 1: <model response 1> \}, \\
        \{response 2: <model response 2> \}, \\
        ... \\
        \{response K: <model response K> \}, \\ ] 
\end{mtbox}

SIMS-CS extends the self-improving steering loop by introducing a contrastive sampling strategy that persistently curates the most contrastive prompt–response pairs encountered. For each iteration~$t$, the current policy~$\pi_{t-1}$ draws $N$ prompts $\{\mathbf{x}_{n}\}_{n=1}^{N}\sim D_{\text{prompt}}$ and generates $K$ candidate responses $\{\mathbf{y}_{n,k}\}_{k=1}^{K}\sim\pi_{t-1}(\,\cdot\mid\mathbf{x}_{n})$. The preference oracle $o$ returns pair-wise probabilities $\mathbb{P}_{o}(\mathbf{y}_{n,k}\succ\mathbf{y}_{n,k'})$, from which we compute a contrastive reward 
\begin{equation}
r_i \;=\;
    \max_{k}\,\mathbb{P}_o\bigl(y_{i,k}\succ\pi_{t}| x_i\bigr)
     \;-\;
     \max_{k}\,\mathbb{P}_o\bigl(y_{i,k}\prec\pi_{t}| x_i\bigr),
\end{equation}
Each triple $(\mathbf{x}_{n},\{\mathbf{y}_{n,k}\}_{k=1}^{K},r_{n})$ is appended to~$\mathcal{B}$.  After processing all prompts we select the \emph{top-$N$} entries of $\mathcal{B}$ by reward to form $\mathcal{D}_{t}=\text{Top-}N(\mathcal{B};r)$. For every $(\mathbf{x},\mathbf{y},r)\in\mathcal{D}_{t}$ we isolate the most-preferred response $\mathbf{y}_{(1)}$ and the least-preferred response $\mathbf{y}_{(K)}$, yielding
\begin{equation}
\mathcal{D}^{+}_{t}=\{(\mathbf{x},\mathbf{y}_{(1)})\},\qquad
  \mathcal{D}^{-}_{t}=\{(\mathbf{x},\mathbf{y}_{(K)})\}.
\end{equation}
Layer-wise activations are collected as $\mathcal{H}^{+}_{l}=\{\mathcal{M}_{l}(\mathbf{x},\mathbf{y})\}_{(\mathbf{x},\mathbf{y})\in\mathcal{D}^{+}_{t}}$ and $\mathcal{H}^{-}_{l}=\{\mathcal{M}_{l}(\mathbf{x},\mathbf{y})\}_{(\mathbf{x},\mathbf{y})\in\mathcal{D}^{-}_{t}}$. The steering-rule learner then solves 
$$
\bigl\{f^{(t)}_{l},f'^{(t)}_{l}\bigr\}_{l=1}^{L}
    \;=\;
    \mathcal{A}\!\bigl(\mathcal{H}^{+}_{1:L},\mathcal{H}^{-}_{1:L}\bigr),
$$
producing transforms that nudge hidden states toward the positive manifold and away from the negative one.

\begin{algorithm}\small
\caption{\textsc{Self-Improving Steering with Contrastive Sampling (SIMS-CS)}}
\label{alg:sims-cs}
\KwIn{Language model $\mathcal{M}$ with $L$ layers;  
      preference oracle $o$; steering-rule learner $\mathcal{A}$;  
      prompt distribution $D_{\!\text{prompt}}$; iterations $T$;  
      prompts per iteration $N$; responses per prompt $K$}
Initialize steering transforms  
      $\{f^{(0)}_{l}\}_{l=1}^{L}$ and $\{f'^{(0)}_{l}\}_{l=1}^{L}$\;
Define initial policy  
      $\pi_{0}=(\mathcal{M},\{f^{(0)}_{l}\}_{l=1}^{L},\{f'^{(0)}_{l}\}_{l=1}^{L})$\;
Initialize global \emph{response bank} $\mathcal{B}\leftarrow \varepsilon$\;
\For{$t=1$ \textbf{to} $T$}{
    Sample prompts $\{\mathbf{x}_{n}\}_{n=1}^{N}\sim D_{\!\text{prompt}}$\;
    \For{$n=1$ \textbf{to} $N$}{
         Generate $K$ candidate responses  
               $\{\mathbf{y}_{n,k}\}_{k=1}^{K}\sim\pi_{t-1}(\,\cdot\mid\mathbf{x}_{n})$\;
        Query oracle $o$ for pairwise preferences  
               $\mathbb{P}_o(\mathbf{y}_{n,k}\succ\mathbf{y}_{n,k'})$\;
        \textbf{Contrastive sampling:} Compute a contrastive scalar reward as $ r_{n} = \max_{k \in K} \mathbb{P}_o(y_{n,k} \succ \bm{\pi}_t \mid x_n) - \max_{k \in K} \mathbb{P}_o(y_{n,k} \prec \bm{\pi}_t \mid x_n) $\;
        \textbf{Contrastive sampling:} Add tuples to bank  
               $\mathcal{B}\leftarrow\mathcal{B}\cup
               (\mathbf{x}_{n},\{\mathbf{y}_{n,k}\}_{k=1}^{K},r_{n})$\;
    }
    \textbf{Contrastive sampling:}  
        $\mathcal{D}_t = \text{Top-}N(\mathcal{B})$\;
    \textbf{Contrastive sampling:}  Construct datasets
        $\mathcal{D}^{+}_{t}=\bigl\{(\mathbf{x}_{n},\mathbf{y}_{n,(1)})\}_{(\mathbf{x},\mathbf{y},r)\in\mathcal{D}_{t}},
          \quad
          \mathcal{D}^{-}_{t}=\bigl\{(\mathbf{x}_{n},\mathbf{y}_{n,(K)})\bigr\}_{(\mathbf{x},\mathbf{y},r)\in\mathcal{D}_{t}}$\;
    Collect hidden activations  
        $\displaystyle
          \mathcal{H}^{+}_{l}=\{\mathcal{M}_{l}(\mathbf{x},\mathbf{y})\}_{(\mathbf{x},\mathbf{y})\in\mathcal{D}^{+}_{t}},\;
          \mathcal{H}^{-}_{l}=\{\mathcal{M}_{l}(\mathbf{x},\mathbf{y})\}_{(\mathbf{x},\mathbf{y})\in\mathcal{D}^{-}_{t}}$\;
    Learn new steering functions  
        $\displaystyle
          \{f^{(t)}_{l},f'^{(t)}_{l}\}_{l=1}^{L}
          =\mathcal{A}\bigl(\mathcal{H}^{+}_{1:L},\mathcal{H}^{-}_{1:L}\bigr)$\;
    Update policy  
          $\pi_{t}=(\mathcal{M},\{f^{(t)}_{l}\}_{l=1}^{L},\{f'^{(t)}_{l}\}_{l=1}^{L})$\;
}
\Return $\pi_{T}$\;
\end{algorithm}

\section{Sample Model Responses}
\label{sec:samples}

\begin{mtbox}{\small \textbf{Prompt}: \{ What are the names of some famous actors that started their careers on Broadway? \}}
  \small

    \textbf{\ttfamily Llama3-8B}: Many famous actors got their start on Broadway before making it big in Hollywood. Here are some examples:
    1.~Julie Andrews – She originated the title role in ``My Fair Lady'' on Broadway in 1956 and won a Tony Award for Best Actress.
    2.~Hugh Jackman – He originated the role of Curly in the Broadway production of ``Oklahoma!'' in 1998 and later won a Tony Award for Best Actor for ``The Boy from Oz.''
    3.~Neil Patrick Harris\\[6pt]

    \textbf{MS}: Many talented actors have, indeed, begun their careers on the Great White Way. Here are a few examples:
    1.~Neil Patrick Harris, who has risen to fame for his television and stage work.
    2.~Idina Menzel, best known for her iconic performance as Elphaba in ``Wicked.''\\[6pt]

    \textbf{SIMS}: Many talented actors have gotten their start on Broadway before transitioning to film and television. Here are a few examples:
    1.~Hugh Jackman: Started his career in Australia with roles like ``Oklahoma!'' and later originated the role of Curly in the 1998 West End revival.
    2.~Neil Patrick Harris\\[6pt]

    \textbf{SIMS-PR}: Many talented actors have gotten their start on Broadway before transitioning to film and television. Here are a few examples:
    1.~Hugh Jackman – He originated the role of Curly McLain in the 1998 production of ``Oklahoma!'' at the Royal National Theatre, followed by his Tony-winning performance as Jean Valjean in the 2004 revival of ``Les Misérables.''
    2.~Neil Warren\\[6pt]

    \textbf{SIMS-CS}: Many talented actors have made a name for themselves in Hollywood after starting out on Broadway. Here are some examples:
    1.~Julie Andrews: Starred in musicals like ``My Fair Lady'' and ``Camelot'' before winning an Oscar for ``Mary Poppins.''
    2.~Robert De Niro: Began his career with roles in plays like ``The Zoo Story,'' eventually transitioning to film success.
    3.~Al Pacino
\end{mtbox}

\begin{mtbox}{\small \textbf{Prompt}: \{\,How did US states get their names?\}}
    \small

    \textbf{\texttt{Mistral-7B}}:~The naming of U.S.\ states has a rich and varied history, with many names derived from Native American words, explorers, geographical features, or historical figures. Here are some common ways U.S.\ states got their names:\\
    1.~\emph{Native American words}: Many states have names derived from Native American languages. For example, \emph{California} is sometimes linked to the fictional island “Calafia,” while \emph{Florida} comes from the Spanish \emph{Pascua Florida}.\\[6pt]

    \textbf{MS}:~The naming of U.S.\ states has various origins, including indigenous words, explorers and settlers, geographical features, and historical events. Here are some of the most common ways that states got their names:\\
    1.~\emph{Indigenous words}: Many state names come from Native American languages. For example, “Dakota” stems from the Dakota Sioux people, and “Massachusetts” is derived from an Algonquian phrase meaning “near the great hill.”\\[6pt]

    \textbf{SIMS}:~The naming of U.S.\ states reflects a rich and diverse history, with sources including Native American words, explorers, geographical features, and founding fathers. Here are some common ways that states got their names:\\
    1.~\emph{Native American names}: Many states have names taken from the languages of the indigenous peoples who inhabited those areas before European settlers arrived. An example is \emph{Mississippi}, from the Ojibwe \emph{Misi-ziibi} (“great river”).\\[6pt]

    \textbf{SIMS-PR}:~The naming of U.S.\ states has a rich and varied history, with many factors influencing the choices made. Here are some common reasons for the names of various U.S.\ states:\\
    1.~\emph{Geographical features}: Several states were named after prominent natural landmarks, such as mountains (\emph{Colorado}, \emph{Nevada}) or forests (\emph{Vermont}, from French \emph{vert mont}, “green mountain”).\\[6pt]

    \textbf{SIMS-CS}:~The naming of U.S.\ states can be traced back to exploration, indigenous peoples, natural features, and political influences. Here are some common reasons behind the names of several U.S.\ states:\\
    1.~\emph{Exploration}: Some states were named to honor explorers or monarchs. For example, \emph{Virginia} honors Queen Elizabeth I (the “Virgin Queen”), and \emph{Louisiana} honors King Louis XIV of France.\\
\end{mtbox}


